\documentclass[letterpaper]{article} 
\usepackage{aaai2026}  
\usepackage{times}  
\usepackage{helvet}  
\usepackage{courier}  
\usepackage[hyphens]{url}  
\usepackage{graphicx} 
\usepackage{natbib}  
\usepackage{caption} 
\usepackage{algorithm}
\usepackage{algorithmic}
\usepackage{amsfonts}
\usepackage{hyperref}
\usepackage{booktabs}
\usepackage{multirow}
\usepackage{amsthm}
\usepackage{newfloat}
\usepackage{listings}
\DeclareCaptionStyle{ruled}{labelfont=normalfont,labelsep=colon,strut=off} 
\floatstyle{ruled}
\newfloat{listing}{tb}{lst}{}
\floatname{listing}{Listing}
\title{GT$^2$-GS: Geometry-aware Texture Transfer for Gaussian Splatting}

\author {
    Wenjie Liu\textsuperscript{\rm 1},
    Zhongliang Liu\textsuperscript{\rm 2},
    Junwei Shu\textsuperscript{\rm 1},
    Changbo Wang\textsuperscript{\rm 3},
    Yang Li\textsuperscript{\rm 1}\equalcontrib
}
\affiliations {
    \textsuperscript{\rm 1}School of Computer Science and Technology, East China Normal University, Shanghai, China\\
    \textsuperscript{\rm 2}School of Software Engineering, East China Normal University, Shanghai, China\\
    \textsuperscript{\rm 3}School of Data Science and Engineering, East China Normal University, Shanghai, China\\
    \{51265901068, 10235101440, 51265901091\}@stu.ecnu.edu.cn, \{cbwang, yli\}@cs.ecnu.edu.cn
}

\usepackage{bibentry}

\begin{document}

\maketitle

\begin{abstract}

Transferring 2D textures onto complex 3D scenes plays a vital role in enhancing the efficiency and controllability of 3D multimedia content creation. However, existing 3D style transfer methods primarily focus on transferring abstract artistic styles to 3D scenes. These methods often overlook the geometric information of the scene, which makes it challenging to achieve high-quality 3D texture transfer results. In this paper, we present GT$^2$-GS, a geometry-aware texture transfer framework for gaussian splatting. First, we propose a geometry-aware texture transfer loss that enables view-consistent texture transfer by leveraging prior view-dependent feature information and texture features augmented with additional geometric parameters. Moreover, an adaptive fine-grained control module is proposed to address the degradation of scene information caused by low-granularity texture features. Finally, a geometry preservation branch is introduced. This branch refines the geometric parameters using additionally bound Gaussian color priors, thereby decoupling the optimization objectives of appearance and geometry. Extensive experiments demonstrate the effectiveness and controllability of our method. Through geometric awareness, our approach achieves texture transfer results that better align with human visual perception. Our homepage is available at \href{https://vpx-ecnu.github.io/GT2-GS-website}{https://vpx-ecnu.github.io/GT2-GS-website}.
\end{abstract}

\section{Introduction}
\label{sec:intro}

3D style transfer~\cite{zhang2022arf,zhang2024stylizedgs} aims to transfer the stylistic elements of a reference image onto a 3D scene while preserving the scene's original structural and semantic information. With the rapid development of fields such as virtual reality, robotics, film, and gaming, the demand for high-quality 3D content has increased significantly. 3D stylization techniques offer a promising solution by accelerating the creation of 3D content, particularly in complex 3D scene environments.

Recently, the emergence of Neural Radiance Fields (NeRF)~\cite{mildenhall2020nerf} and 3D Gaussian Splatting (3DGS)~\cite{kerbl3Dgaussians} has significantly advanced the field of 3D stylization. NeRF-based stylization methods~\cite{zhang2022arf, zhang2023ref, nguyen2022snerf}, leveraging the advantages of implicit representations, allow for the decoupled optimization of appearance and geometry. In contrast, 3DGS-based approaches~\cite{galerne2025sgsst,liu2025abc, liu2024stylegaussian} offer benefits such as explicit editability and real-time rendering. 3D stylization methods can be broadly categorized into feed-forward and optimization-based approaches, depending on whether they support zero-shot style transfer. While feed-forward methods~\cite{liu2023stylerf, huang2022stylizednerf, liu2024stylegaussian} enable zero-shot stylization, they often exhibit lower rendering quality. In contrast, optimization-based methods~\cite{zhang2022arf, liu2025abc,zhang2024stylizedgs,galerne2025sgsst} typically produce higher-fidelity stylization results. Despite these differences, both types of methods commonly define optimization objectives within the VGG feature space during training.

\begin{figure}[t]
    \centering
    \includegraphics[width=1\linewidth]{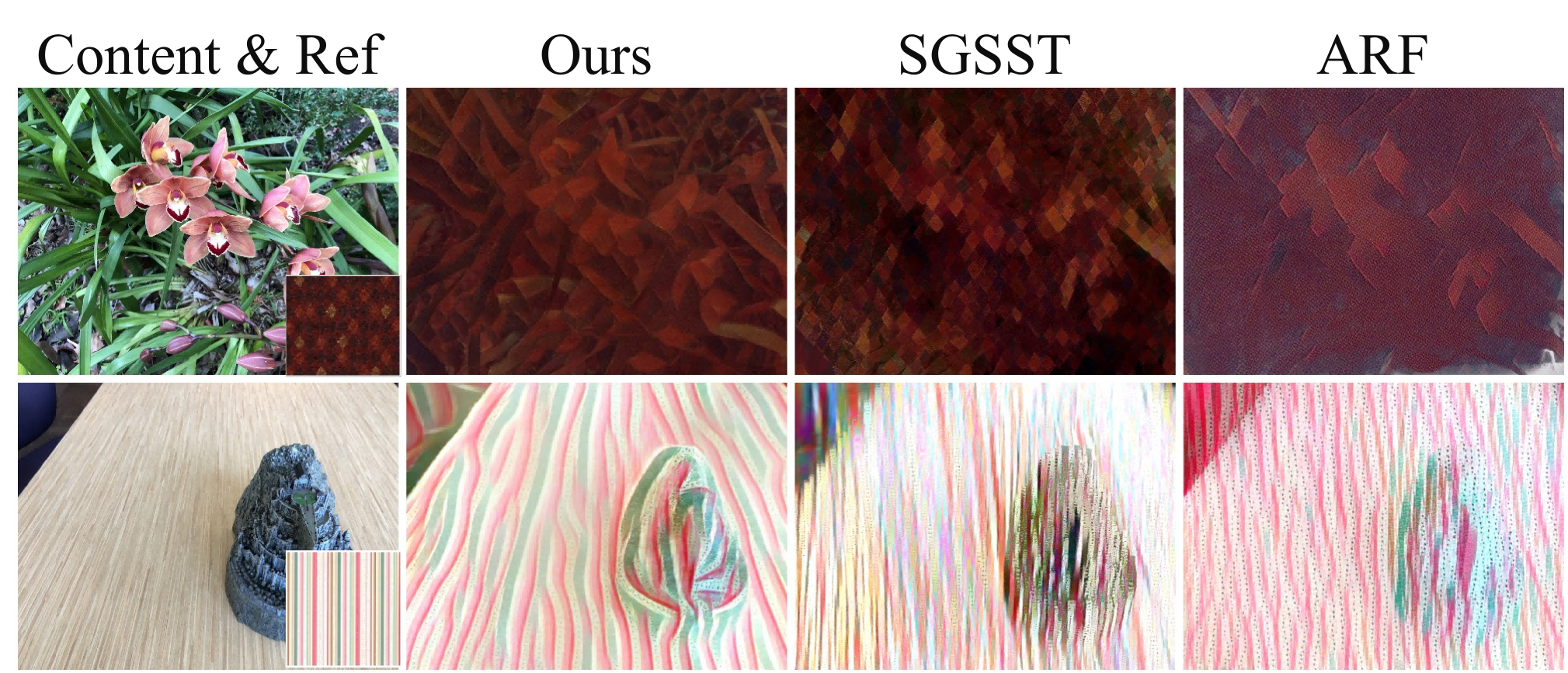}
    \caption{Previous state-of-the-art methods struggle to accurately transfer complex textures to 3D scene. In contrast, our GT$^2$-GS framework incorporates geometric information into the optimization process, enabling geometry-aware texture transfer.}
    \label{fig:teaser}
\end{figure}

Texture is a crucial component of various graphics pipelines and plays a vital role in producing high-quality 3D assets. Compared to abstract artistic styles, texture elements are more controllable for users. However, as illustrated in Fig.~\ref{fig:teaser}, existing methods struggle to accurately transfer the texture of the reference image onto the 3D scene. We analyze the limitations of current approaches from the perspective of the scene optimization process. First, during scene optimization, multi-view objectives should preserve correct geometric relationships. For example, using content images captured from the same scene as ground truth. However, existing style transfer methods define their objectives independently across views, without considering the rich geometric structure within the scene or the geometric consistency across viewpoints. Moreover, we observe a fine-grained mismatch between the VGG feature space and the pixel space. Commonly used style transfer losses fail to account for this mismatch, which leads to regions with high pixel-level information density being easily disrupted by coarse-grained feature representations. Addressing these issues is crucial for achieving high-quality texture transfer.

In this paper, we propose a novel framework GT$^2$-GS for achieving geometry-aware texture transfer. Our method explicitly accounts for geometric information and fine-grained discrepancies, enabling texture transfer results that better align with human visual perception. We first propose a geometry-aware texture transfer loss, which is built upon texture features augmented with additional geometric parameters and incorporates cross-view geometric information to ensure consistent and controllable texture transfer. In addition, an adaptive fine-grained control module is proposed. It adaptively adjusts the strength of texture learning based on the information density of different pixel regions, thereby mitigating the fine-grained discrepancy between texture features and pixel-level representations. To address the coupling between Gaussian geometry and color parameters, we further introduce a geometry preservation branch. We bind additional color parameters to the Gaussians and optimize them using the content image as ground truth, in order to accurately refine the scene geometry. Extensive experiments demonstrate that our proposed framework achieves high-quality texture transfer results.

Our main innovations are as follows:
\begin{itemize}
\item We propose a geometry-aware texture transfer framework for general 3DGS scenes. The proposed framework enables controllable and high-quality texture transfer results.

\item Our Geometry-aware Texture Transfer Loss and Adaptive Fine-Grained Control Module respectively account for geometric information and fine-grained discrepancies between texture features and pixels, enabling high-quality texture transfer. Moreover, the Geometry Preservation Branch provides a novel approach to preserving geometry during appearance editing.
\item Extensive experiments demonstrate that, compared to SOTA methods, our proposed work achieves texture transfer results more consistent with human visual perception. Our proposed method maintains real-time rendering and multi-view consistency. The source code will be released. 
\end{itemize}

\section{Related Work}
\subsection{Texture Transfer}

Early texture transfer algorithms~\cite{ashikhmin2003fast, lee2010directional, hertzmann2023image, efros2023image} were based on non-parametric texture synthesis. These methods preserve the structural information of the target image to achieve texture transfer effects. Gatys et al.~\cite{gatys2016image} pioneered the use of convolutional neural networks (CNNs)~\cite{simonyan2014very} for image style transfer, achieving impressive results. Subsequently, a variety of deep learning-based style transfer methods~\cite{chen2016fast,li2016combining,huang2017arbitrary,huo2021manifold} have been proposed. These methods are capable of transferring abstract artistic elements such as color distribution, brushstroke characteristics, and overall composition, rather than focusing solely on texture transfer. 
Leveraging the powerful feature extraction capabilities of networks such as CNNs and Transformers~\cite{vaswani2017attention}, some deep learning-based texture transfer methods~\cite{wang2022texture, chen2022auv,pu2024dynamic} can transfer local textures from the reference image to the target image based on high-level semantic correspondence. In this work, we explore how to directly transfer texture features into 3D representations.

\begin{figure*}[h]
    \centering
    \includegraphics[width=0.9\linewidth]{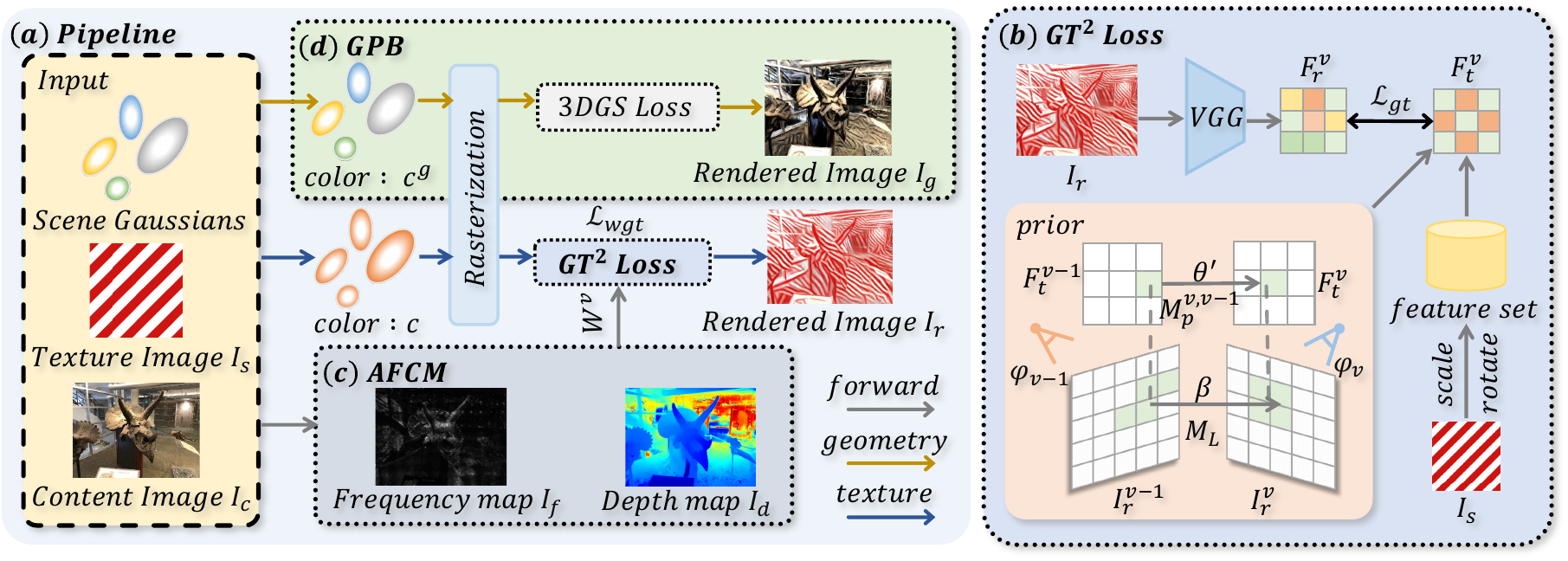}
    \caption{Overview of GT$^2$-GS. The overall pipeline is illustrated in (a). The input to our proposed framework includes the scene Gaussians, content images, and a texture image. The texture image is transformed into a texture feature set, which is used to construct the target feature map in the GT$^2$ Loss, as shown in (b). We extract pixel-wise information from the input and transform it into an adaptive weight matrix to control texture learning, as illustrated in (c). After embedding the parameter $c^g$, the Gaussians are optimized using two branches. The additional geometry preservation branch is shown in (d). As a result, the texture features are integrated into the Gaussian representation.}
    \label{fig:pipeline}
\end{figure*}

\subsection{3D Style Transfer}

3D style transfer aims to transfer the style from 2D images to the appearance of 3D scenes, while maintaining the scene's content and multi-view consistency. Early research in 3D scene style transfer focused on explicit representations such as point clouds~\cite{cao2020psnet,huang2021learning}, meshes~\cite{hollein2022stylemesh,yin20213dstylenet}, and voxels~\cite{guo2021volumetric,klehm2014property}. Zhang et al.~\cite{zhang2022arf} explored the potential of NeRF in 3D stylization tasks. By optimizing the scene representation with the proposed 3D style transfer loss, it achieved pleasing visual effects. Subsequently, a variety of NeRF-based 3D style transfer methods~\cite{huang2022stylizednerf,pang2023locally,liu2023stylerf,zhang2024coarf,jung2024geometry} have emerged. Recently, the emergence of 3DGS~\cite{kerbl3Dgaussians} has brought new possibilities for scene stylization. It features fast training speed, high rendering quality, and efficient performance. Some works~\cite{liu2024stylegaussian,zhang2024stylizedgs,mei2024regs,liu2025abc} have already explored the potential of stylization under this representation. Currently, 3DGS-based scene stylization methods can be categorized into feed-forward approaches~\cite{liu2024stylegaussian} and optimization-based approaches~\cite{zhang2024stylizedgs,mei2024regs,liu2025abc}. Feed-forward methods enable zero-shot style transfer. For instance, StyleGaussian~\cite{liu2024stylegaussian} embeds VGG features into Gaussians and leverages pre-trained scene representations combined with AdaIN~\cite{huang2017arbitrary} to achieve real-time stylization. On the other hand, optimization-based methods attain higher-quality style transfer results. Some works~\cite{zhang2024stylizedgs,liu2025abc} further exploit the explicit characteristic of Gaussian representations to enable region-controllable scene stylization.
However, existing methods do not incorporate geometric information during the stylization process, making them unsuitable for texture images. Meanwhile, these methods overlook the fine-grained discrepancies between features and pixels. By incorporating geometric information and adaptive fine-grained control, our method is able to achieve high-quality texture transfer results.

\section{Preliminaries}
\subsection{3D Gaussian Splatting}
3D Gaussian Splatting~\cite{kerbl3Dgaussians} is a novel explicit 3D representation method. It uses a set of parameterized Gaussians  $G=\{g_i\}$ to represent 3D scenes. The parameters of each Gaussian $g_i$ include a mean vector $\mu_i$ representing its center position, a covariance matrix $\Sigma_i$ describing its shape, an opacity parameter $\sigma_i$ and a color parameter $c_i$, which is represented as spherical harmonic coefficients. Among these, the covariance matrix $\Sigma_i$ is decomposed into rotation parameters $r_i$ and scaling parameters $s_i$ to effectively maintain its positive semi-definite property during the optimization process. The optimization parameters of Gaussian $g_i$ are actually represented as $g_i=\{\mu_i, r_i, s_i, \sigma_i, c_i\}$. The 3D Gaussians are efficiently rendered through a fast differentiable rasterization~\cite{lassner2021pulsar}. Specifically, the Gaussians are grouped into tiles and sorted by depth. The color $C$ of a pixel is computed through $\alpha$-blending:
\begin{equation}
    C = \sum_{i \in N} T_i \alpha_i c_i, T_i=\prod_{j=1}^{i-1} (1 - \alpha_j),
\end{equation}
where $T_i$ is the transmittance and $\alpha_i$ is the alpha-compositing weight for the $i$-th Gaussian. The depth value $D$ can be obtained using a method similar to rendering color,
\begin{equation}
    D = \sum_{i \in N} T_i \alpha_i d_i,
    \label{Equation:depth}
\end{equation}
where $d_i$ is the depth for $i$-th Gaussian. During optimization, 3DGS adopts an adaptive densification strategy to control the distribution of Gaussians. Specifically, this strategy performs cloning or splitting of Gaussians based on their positional gradients and sizes.

\subsection{Style Transfer Loss}
By computing the loss on the feature maps of rendered views, the learnable parameters in the scene representation can be optimized. Existing optimization-based 3D style transfer methods~\cite{zhang2022arf,pang2023locally,zhang2024coarf,zhang2024stylizedgs} typically adopt the nearest neighbor feature matching (NNFM) loss~\cite{zhang2022arf} as their style loss function. It matches the rendered features with the nearest neighbor features in the style feature set and minimizes the cosine distance between them. Specifically, a random viewpoint is selected to obtain the rendered image $I_r$. The same feature extractor (e.g., VGG~\cite{simonyan2014very}) is used to extract the feature maps $F_r$ and $F_s$ from $I_r$ and the style image $I_s$, respectively. Let $F_r(i,j)$ denote the feature vector at the pixel location $(i,j)$ of the rendered feature map $F_r$, the NNFM loss can be expressed as

\begin{equation}
    L_{style} = \frac{1}{N} \sum_{i,j} \min_{i',j'} dist(F_r(i,j),F_s(i',j')),
    \label{Equation:nnfm}
\end{equation}
where $N$ is the number of pixels in $F_r$, $dist(a,b)$ is the cosine distance between two feature vectors $a$ and $b$:
\begin{equation}
    dist(a,b) = 1 - \frac{a \cdot b}{\|a\| \|b\|}.
\end{equation}

\section{Method}
\subsection{GT$^2$-GS Framework Overview}

In this section, we provide a detailed overview of the proposed GT$^2$-GS framework. Geometry-aware Texture Transfer Loss (GT$^2$ Loss) is proposed to enable controllable and high-quality texture transfer. It is computed using texture features bound with geometric parameters and the prior information from the previous viewpoints. 
Furthermore, we propose the Adaptive Fine-grained Control Module (AFCM) to regulate the texture learning intensity based on the scene information density in the pixel space. This prevents excessive information from being corrupted by low-granularity features. Finally, for the input Gaussian, an additional color parameter $c^g$ is embedded and initialized using the current color parameter $c$. The Geometry Preservation Branch (GPB) is introduced based on $c^g$, with the aim of decoupling the optimization processes of the geometry and appearance of the scene. The pipeline is shown in Fig. \ref{fig:pipeline}.

\begin{figure}[t]
    \centering
    \includegraphics[width=1\linewidth]{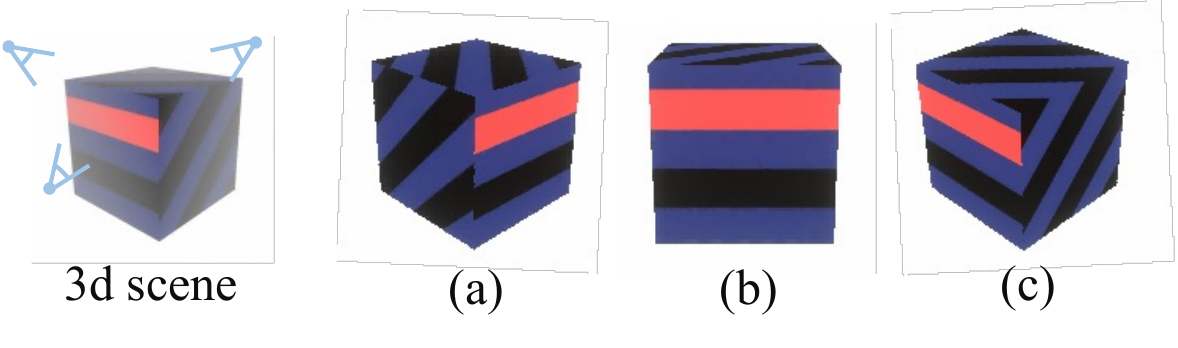}
    \caption{Differences in Perspective. The same textured region (highlighted in red) on the 3D object exhibits different texture orientations under varying viewpoints.}
    \label{fig:svd}
\end{figure}

\subsection{Geometry-aware Texture Transfer Loss}
Existing methods primarily focus on artistic style transfer and are mostly based on NNFM loss. As shown in Eq.~\ref{Equation:nnfm}, the objective considers only the relationship between the rendered feature map and the style feature map. The optimization objectives across different viewpoints are constructed independently, without accounting for the underlying geometric relationships between views. However, texture is inherently tied to the geometry of the scene. To achieve accurate texture transfer, we incorporate geometric information into the loss function computation during the optimization process.

First, we associate geometric parameters with each texture feature. Considering the perspective geometry within the scene, edge geometric structures, and transformations across viewpoints, we apply scaling and rotation operations to the texture images. Texture features are then extracted from the transformed images and aggregated into a feature set. Specifically, we obtain the original scene depth map using Eq. \ref{Equation:depth}. The depth values are then sorted and discretized into $K$ groups based on predefined depth intervals. The scaling factor for each group is computed as the ratio between the lowest group depth value $Z_1$ and the group’s depth value $Z_k$. Next, each scaled image is rotated by an angle $\theta$ to capture multi-directional information present in rendered views. Each feature in the texture feature set is denoted as $\{f_{k,\theta}\}$. The scaling parameter $k$ and rotation angle $\theta$ are retained for subsequent computations.

Based on the constructed set of texture features, we propose the GT$^2$ Loss to enable geometry-consistent texture transfer. We construct a per-pixel matched target feature map $F_t$ for the rendered feature map $F_r$. Texture transfer is achieved by minimizing the cosine similarity between the corresponding feature vectors of $F_r$ and $F_t$. 
To ensure that the feature maps $F_t$ constructed from multiple viewpoints maintain correct geometric relationships, we incorporate the construction result of the previous viewpoint’s feature map 
$F_t^{v-1}$ when building the target feature map 
$F_t^v$ for the current viewpoint. The computation process of the homography matrix $M_p^{v,v-1}$ between two viewpoints is formulated as
\begin{equation}
    M_p^{v,v-1} = K_{v-1}[R_{v-1}|T_{v-1}]{[R_v|T_v]^{-1}}{K_v^{-1}},
\end{equation}
where $K_v$ and $[R_v|T_v]$ represent the intrinsic matrix and world-to-camera extrinsic parameters of the $v$-th viewpoint, respectively. Through the homography matrix, the current viewpoint can sample the prior feature $f_{k',\theta'}$ in the screen coordinate system of the prior viewpoint. Considering occlusion relationships, we filter the projected points using the depth map $I_d^{v-1}$ of the prior viewpoint.

However, as shown in Fig.~\ref{fig:svd}, the orientation of the same texture region varies across different viewing angles. Each texture feature vector inherently contains scale and orientation. Directly using this feature vector as a prior fails to account for the impact of viewpoint changes in 3D space. To address this, we utilize upsampling to obtain the pixel set $\{p_v\}$ corresponding to each feature map location in the pixel coordinate system. Through the transformation relationship $M_p^{v-1,v}$, we can determine its corresponding pixel set $\{p_{v-1}\}$ in the previous viewpoint. We use the least squares method to compute the linear transformation matrix $M_L \in \mathbb{R}^{2\times2}$ for texture variation between viewpoints. The obtained transformation matrix $M_L$ can be decomposed using SVD to extract the rotation angle $\beta$. The construction method of the texture feature vector at position $(i,j)$ in the target feature map $F_t$ is formulated as

\begin{equation}
    F_t(i,j)=\mathop{\arg\min}\limits_{f_{k,\theta}} 
    dist(F_r(i,j),f_{k,\theta})+\lambda_p|\theta'+\beta-\theta|,
    \label{Equation:target}
\end{equation}

where $\lambda_p$ is the prior texture orientation control coefficient. Subsequently, texture transfer can be achieved by minimizing the cosine similarity between the feature vectors at corresponding positions in the rendered feature map $F_r^v$ and the target feature map $F_t^v$ for the current viewpoint $v$. The geometry-aware texture transfer loss function is formulated as follows:

\begin{equation}
    L_{gt} = \frac{1}{N} \sum_{i,j} dist(F_r^v(i, j), F_t^v(i,j)).
    \label{Equation:gt}
\end{equation}

\subsection{Adaptive Fine-grained Control Module}
Most existing optimization-based style transfer methods rely on VGG features. However, feature maps extracted through multiple layers of convolutional neural networks exhibit significantly lower granularity compared to the original image pixels. When optimizing the scene using only texture transfer losses, this can lead to the following issues: First, due to perspective projection, regions with greater depth tend to concentrate more scene information. Learning coarse-grained texture features may overwrite these important details. Second, scenes often contain geometrically fine-grained structures such as stairs and railings, which can be degraded or lost under coarse texture representations. To address these challenges, we propose an Adaptive Fine-Grained Control Module.

Specifically, we first obtain the depth map $I_d$ and frequency density map $I_f$ for each viewpoint from the original Gaussian scene and content image, respectively. Both are rescaled to match the spatial dimensions of the feature maps. Furthermore, we introduce a geometric distortion map $\Phi$, defined as the discrepancy in geometric information between the learned texture features and rendered features. The geometric distortion map is computed by measuring the angular difference between the features obtained using Eq.~\ref{Equation:target} and those derived without any prior information. The adaptive fine-grained control map is formulated as 
\begin{equation}
    W^v = \lambda_{d}(1-I_d^v)+\lambda_{f}(1-I_f^v)+\lambda_{\Phi}(1-\Phi),
\end{equation}
where $I_d^v$, $I_f^v$ and $\Phi$ are all normalized, and $\lambda_{d}$, $\lambda_{f}$ and $\lambda_{\Phi}$ are weight coefficients. In most cases, we aim to alter the appearance of foreground objects while simultaneously satisfying requirements for both shallow depth and high frequency. We therefore employ an additive formulation to balance the relationship between depth and frequency. Furthermore, we regularize the learning process to favor textures with low distortion, thereby better preserving geometric fidelity during texture adaptation. 

The derived adaptive weight matrix $W^v$ is subsequently applied to Eq. \ref{Equation:gt} as
\begin{equation}
    L_{wgt} = \frac{1}{N} \sum_{i,j} W^v(i,j)\ dist(F_r^v(i, j), F_t^v(i,j)),
    \label{Equation:wgt}
\end{equation}
where $L_{wgt}$ is weighted GT$^2$ Loss. The total loss during the texture transfer phase is expressed as
\begin{equation}
    L_{tot} = {\lambda_{wgt}}L_{wgt} + {\lambda_c}L_{content} + {\lambda_{tv}}L_{tv},
    \label{Equation:tot}
\end{equation}
where $L_{tv}$ is the total variation loss, $\lambda_{wgt}$, $\lambda_c$, $\lambda_{tv}$ are the coefficients of the corresponding loss functions.

\begin{figure}[t]
    \centering
    \includegraphics[width=1\linewidth]{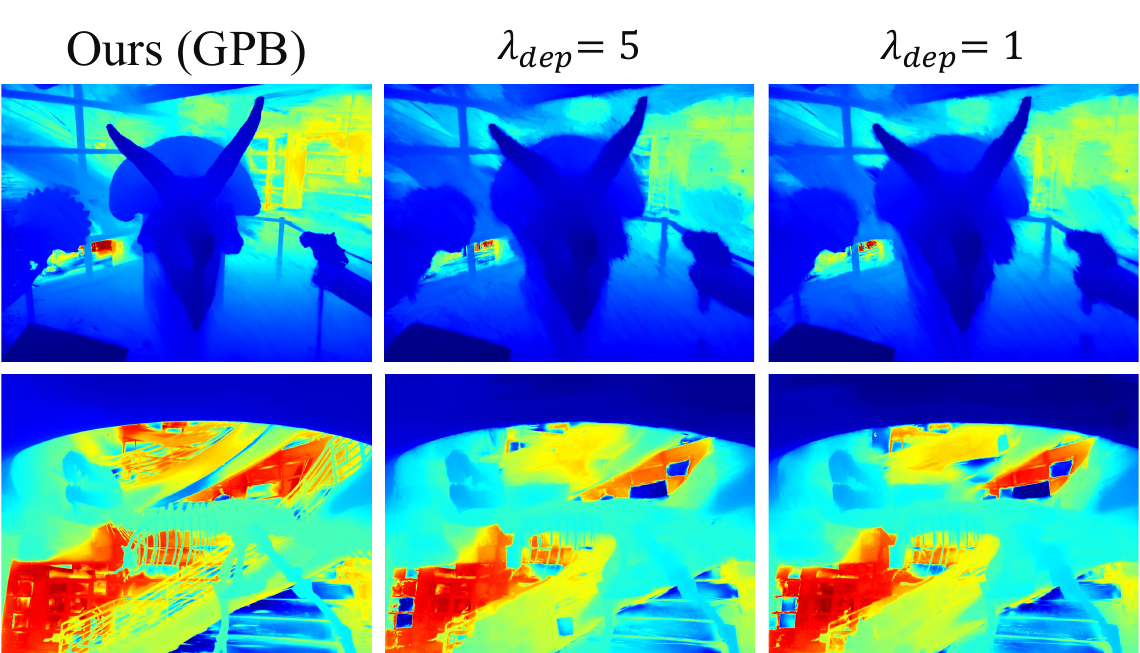}
    \caption{Comparison of Geometry Preservation Results. It can be observed that as the number of Gaussians increases, depth regularization struggles to preserve the geometric information of the scene. $\lambda_{dep}$ is the weighting coefficient for the depth regularization term.}
    \label{fig:depth}
\end{figure}

\begin{figure*}[t]
    \centering
    \includegraphics[width=1\linewidth]{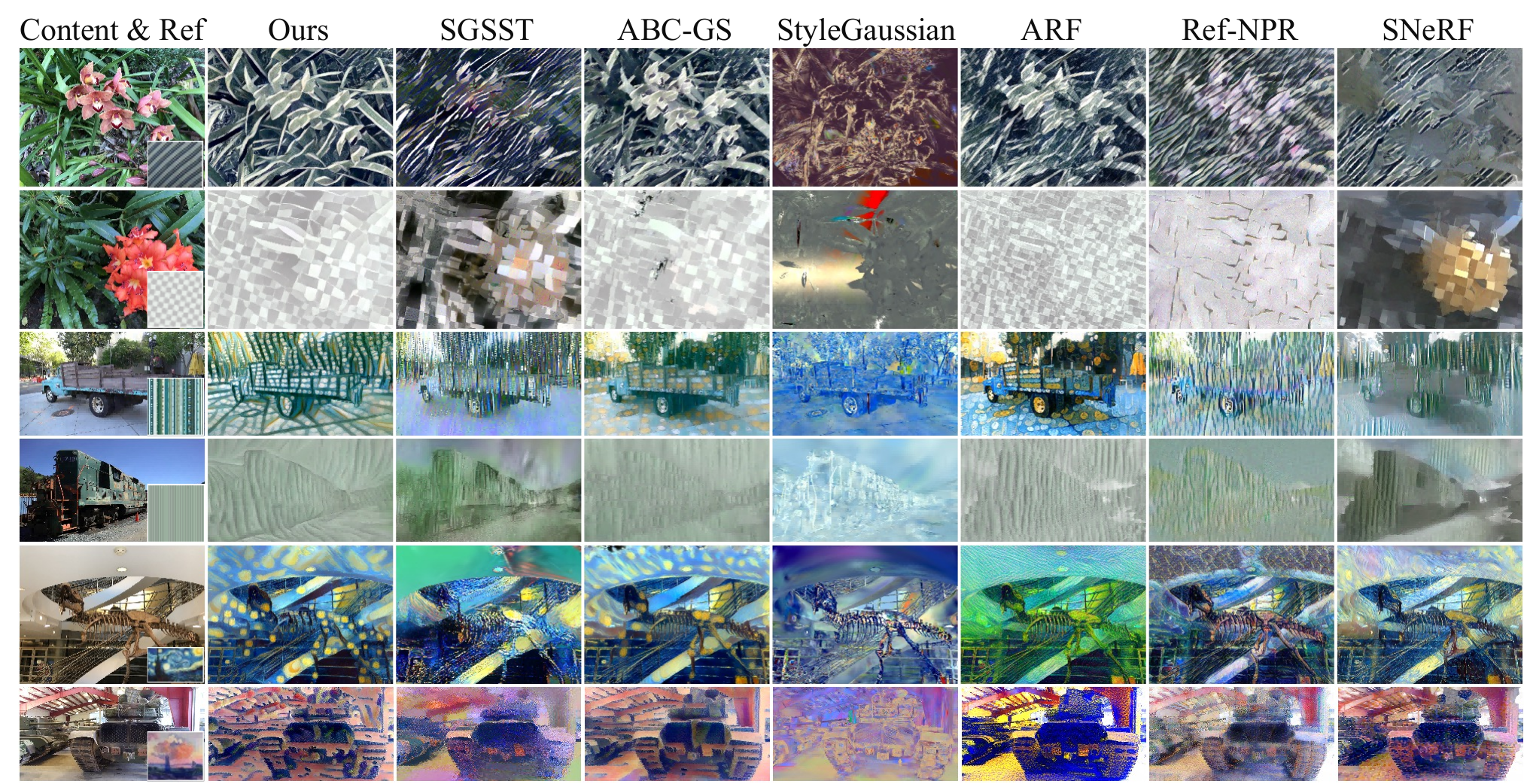}
    \caption{Qualitative Comparison of Texture Transfer and Style Transfer. The first four rows in the figure show the texture transfer results, while the last two rows present the style transfer results.}
    \label{fig:qual_texture}
\end{figure*}

\subsection{Geometry Preservation Branch}
3DGS is an explicit representation, where the scene’s ability to learn textures is closely related to the distribution of Gaussians. In low-texture regions, the density of Gaussians is typically lower. While these regions achieve good rendering quality in the original rendering process, they struggle to learn new texture appearances. The Gaussian densification strategy increases the number of Gaussians in low-texture regions during the texture migration phase, enabling the learning of complex textures. However, Gaussians are densified solely based on gradients. Since texture transfer lacks ground truth during optimization, the densification strategy may introduce erroneous Gaussians floating in space. As shown in Fig. \ref{fig:depth}, simply adding depth regularization does not solve this problem. Unlike NeRF, explicit 3D Gaussians jointly encode both geometric and color parameters. To address this issue, we propose a specialized branch for optimizing the geometric parameters of Gaussians.

Our key insight is to introduce an additional optimization objective focused on geometry preservation during training, in order to balance appearance optimization with geometric integrity. Specifically, we associate each Gaussian with an additional color parameter $c^g$, initialized using the original color values from the scene. During optimization, we render an image $I_{g}$ using these color parameters and optimize the Gaussian parameters by treating the content image $I_c$ as ground truth. The 3D Gaussian Splatting reconstruction loss function is formulated as
\begin{equation}
    \mathcal{L}_{rec} = (1 - \lambda) \mathcal{L}_1 + \lambda \mathcal{L}_{D-SSIM}, 
    \label{rec_loss}
\end{equation}
where $\mathcal{L}_1$ and $\mathcal{L}_{D-SSIM}$ are calculated between the rendered image $I_{g}$ and the content image $I_{c}$. Through an optimization process with ground truth, the Gaussians in the scene are moved to their correct geometric positions.

\section{Experiment}

\noindent\textbf{Datasets.} For the scene datasets, we utilize the LLFF dataset~\cite{mildenhall2019local} and the Tanks and Temples (T\&T) dataset~\cite{knapitsch2017tanks}, which are collected from the real world. Additionally, we use images from the ARF~\cite{zhang2022arf} style dataset and the DTD dataset~\cite{cimpoi2014describing} as reference image datasets.

\noindent\textbf{Baseline.} We compare our method with the state-of-the-art 3D stylization methods, including SGSST~\cite{galerne2025sgsst}, ABC-GS~\cite{liu2025abc}, StyleGaussian~\cite{liu2024stylegaussian}, ARF~\cite{zhang2022arf}, Ref-NPR~\cite{zhang2023ref}, and SNeRF~\cite{nguyen2022snerf}.
Specifically, SGSST, ABC-GS, and StyleGaussian are based on 3DGS, and ARF, Ref-NPR, and SNeRF are based on NeRF. StyleGaussian is a feed-forward-based method, while others are optimization-based methods.

\subsection{Implementation Details}
We perform view-consistent color transfer~\cite{zhang2022arf} on the rendered images before and after texture transfer and use them as content images to optimize the Gaussian parameters. For the VGG~\cite{simonyan2014very} feature extractor, we employ the conv3 block of VGG-16. 
When associating color parameters with texture features, we set the default number of depth groups $K$ to 4, and the rotation angle $\theta$ is sampled across the full 360 degrees. The hyperparameter for AFCM is denoted as $\{\lambda_{d},\lambda_f,\lambda_{\Phi}\}$ = \{0.8, 0.8, 0.25\}. For texture transfer optimization, we set $\{\lambda_{wgt},\lambda_c,\lambda_{tv}\}$ = \{2, 0.005, 0.02\}. All our experiments are conducted on a single NVIDIA RTX 4090 GPU.

\subsection{Qualitative Evaluation}
To comprehensively evaluate the effectiveness of our method, the qualitative experiments are divided into two parts: texture transfer and style transfer. The results are shown in Fig. \ref{fig:qual_texture}.

\noindent\textbf{Texture Transfer.} 
 Fig.~\ref{fig:qual_texture} shows the texture transfer results of our method compared with other methods. Visually, our results exhibit higher fidelity to the reference texture and better alignment with human visual perception. For example, in the geometrically complex orchids scene (row 1 in Fig.~\ref{fig:qual_texture}), our method produces transfer results that appear as if the texture is wrapped naturally around the surface of the orchids. StyleGaussian, as a zero-shot style transfer method, struggles to handle such complex texture transfers. The results of other optimization-based methods exhibit texture discontinuities and appear noticeably blurry, significantly deteriorating overall visual quality. This is primarily due to the inherent coupling between texture and geometry, which these methods fail to consider during optimization. In contrast, our geometry-aware approach ensures consistent texture transfer across multiple views and coherent 3D results.

\noindent\textbf{Style Transfer.} 
To validate the generalizability of our method, we selected artistic style images as reference inputs and conducted a qualitative comparison of style transfer with other methods. As shown in Fig. \ref{fig:qual_texture}, our approach is more effective in transferring the texture elements of the style image to the 3D scene.
SGSST and ABC-GS disable the Gaussian densification strategy during style transfer optimization, making it challenging to capture rich and detailed style features. For instance, in the trex scene (second-to-last row of Fig.~\ref{fig:qual_texture}), the appearance of the white wall region fails to be effectively optimized. In contrast, our method benefits from the joint effect of the AFCM and GPB. This design enables the model to preserve scene geometry while effectively learning style textures from the reference image. NeRF-based methods tend to capture more abstract stylistic elements, such as brush strokes.

\begin{table*}[t]
\centering
\begin{tabular}{c c c c c c c}
\toprule
\textbf{Methods} & \textbf{SSIM($\uparrow$)} & \textbf{CLIP-score($\uparrow$)} & \textbf{ST-LPIPS($\downarrow$)} & \textbf{ST-RMSE($\downarrow$)} & \textbf{LT-LPIPS($\downarrow$)} & \textbf{LT-RMSE($\downarrow$)} \\
\midrule
Ours & \underline{0.51} & \textbf{0.47} & \underline{0.054} & \underline{0.048} & \underline{0.087} & \underline{0.077} \\
SGSST & 0.45 & 0.44 & 0.075 & 0.072 & 0.119 & 0.108 \\
ABC-GS & \textbf{0.56} & \underline{0.46} & \textbf{0.049} & \textbf{0.041} & \textbf{0.080} & \textbf{0.068} \\
StyleGaussian & 0.41 & 0.40 & 0.058 & 0.052 & 0.097 & 0.082 \\
\midrule
ARF & 0.37 & 0.45 & 0.109 & 0.072 & 0.152 & 0.108 \\
Ref-NPR & 0.35 & 0.42 & 0.092 & 0.069 & 0.137 & 0.102 \\
SNeRF & 0.48 & 0.36 & 0.075 & 0.057 & 0.127 & 0.090 \\
\bottomrule
\end{tabular}
\caption{Quantitative Experiment on Multi-view Consistency and Content Preservation.}
\label{tab:consistency_transposed}
\end{table*}

\subsection{Quantitative Evaluation}
In the 3D scene appearance editing task, maintaining multi-view consistency and preserving scene content information are crucial. To this end, we conducted extensive quantitative experiments from both aspects. We randomly selected 100 scene–reference image pairs to quantitatively evaluate our method.

\noindent\textbf{Multi-view Consistency.}
We evaluate multi-view consistency~\cite{liu2024stylegaussian} using both short-term and long-term consistency metrics. The results are shown in Tab. \ref{tab:consistency_transposed}. It can be observed that both our method and ABC-GS achieve high-quality multi-view consistency. Notably, ABC-GS disables the Gaussian densification strategy during optimization. In contrast, our method maintains multi-view consistency even after applying the densification strategy, demonstrating the effectiveness of the proposed GPB.

\noindent\textbf{Content Preservation.}
For 3D texture transfer, it is essential to ensure that the original scene content remains recognizable while editing the scene's appearance. We use SSIM~\cite{wang2004image} and CLIP-score~\cite{radford2021learning} to assess the preservation of content information. Specifically, SSIM evaluates the structural and informational similarity between two images at the pixel level. In contrast, CLIP-score measures the semantic similarity by computing the cosine similarity between the CLIP embeddings of the two images. As shown in Tab. \ref{tab:consistency_transposed}, our proposed method achieves significantly higher scores on both evaluation metrics. In particular, our method outperforms previous approaches in terms of the CLIP-score. It indicates that incorporating geometric information enables accurate texture transfer to the scene appearance while preserving the semantic content of the scene.

\begin{figure}[t]
    \centering
    \includegraphics[width=1\linewidth]{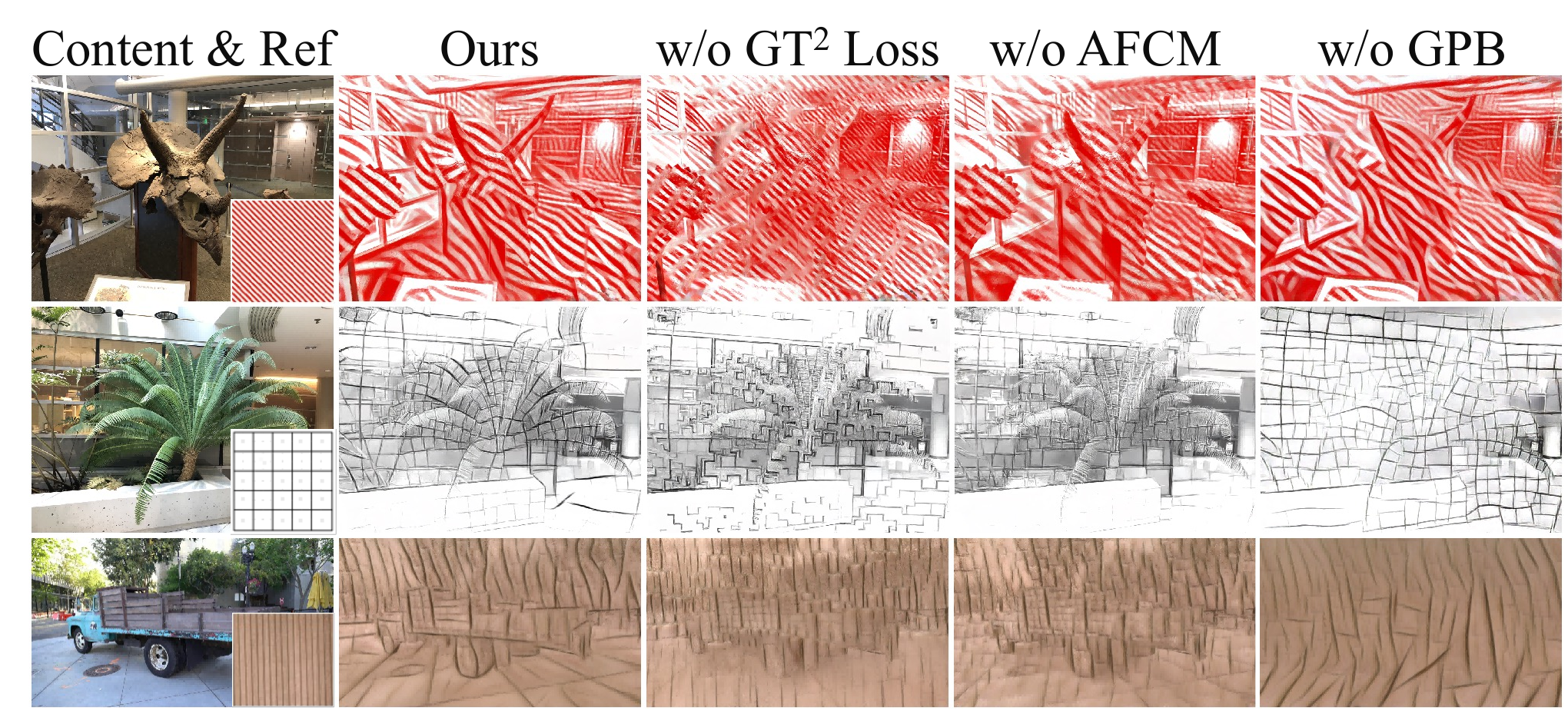}
    \caption{Ablation study for GT$^2$ Loss, AFCM, and GPB.}
    \label{fig:ablation}
\end{figure}

\subsection{Ablation Study}
\noindent\textbf{Impact of GT$^2$ Loss.}
GT$^2$ Loss is introduced to incorporate geometric information into the optimization objective of texture transfer. As shown in Fig.~\ref{fig:ablation}, removing the GT$^2$ Loss results in noticeable texture discontinuities and blurring in the texture transfer outputs. Tab.~\ref{tab:content} further demonstrates the significance of GT$^2$ Loss from a quantitative perspective.

\noindent\textbf{Impact of AFCM.}
The AFCM is introduced to address the granularity mismatch between texture features and pixel-level details. It encourages texture learning in foreground and low-frequency regions, while preserving scene content in background and high-frequency areas. As illustrated in Fig.~\ref{fig:ablation}, in the fern scene (second row), low-texture foreground regions fail to capture style patterns without the presence of AFCM. In the truck scene (third row), which exhibits significant depth variation as a 360° environment, removing AFCM leads to noticeable degradation of geometric and appearance fidelity. 

\noindent\textbf{Impact of GPB.}
The 3D texture transfer tasks inherently lack ground truth supervision, which can introduce incorrect geometry during the scene appearance editing process. To address this, we propose GPB, which leverages the original content images to optimize the geometric parameters of Gaussians and correct inaccurate geometry. As illustrated in Fig. \ref{fig:ablation}, removing GPB results in noticeable artifacts across the scene. Tab.~\ref{tab:content} highlights the importance of GPB in maintaining content fidelity. Through GPB, our texture transfer framework achieves a balance between learning texture features and preserving scene geometry.

\begin{table}[t]
\centering
\setlength{\tabcolsep}{0.8mm}
\begin{tabular}{ccccc}
\toprule
& Ours & w/o GT$^2$ Loss & w/o AFCM & w/o GPB\\
\midrule
SSIM($\uparrow$) & \underline{0.41} & 0.38 & \textbf{0.45} & 0.31\\
CLIP-score($\uparrow$) & \textbf{0.39} & 0.36 & \underline{0.38} &0.37 \\

\bottomrule
\end{tabular}
\caption{Quantitative Ablation Study on Content Protection. The results are obtained from 25 randomly selected experiments conducted on LLFF scenes.}
\label{tab:content}
\end{table}
\section{Conclusion}
In this paper, we introduced GT$^2$-GS, a novel Geometry-aware Texture Transfer framework for Gaussian Splatting that achieves high-quality texture transfer results. Unlike the previous 3D stylization methods, our approach explicitly considered the intrinsic relationship between geometry and texture. We first propose the GT$^2$ Loss, which leverages features augmented with geometric parameters and utilizes cross-view priors to guide scene optimization, enabling geometrically consistent texture transfer. AFCM addresses the granularity mismatch between features and pixels by adaptively controlling the strength of texture learning.
Additionally, GPB introduces a geometry optimization objective grounded in ground-truth appearance, effectively preserving scene geometry during complex stylization. Extensive quantitative and qualitative experiments demonstrated the effectiveness of our proposed method. Moreover, our framework was capable of generating high-quality stylization results, showcasing its generalizability.

As a limitation, because optimizing rendered VGG features requires minimizing texture cosine distance while preserving content loss, which results in a texture interpolated between scene and texture geometry.

\section{Acknowledgments}
This work was supported by the National Natural Science Foundation of China (62572194, 62472178, 62376244) and the Fundamental Research Funds for the Central Universities. This work was supported by the Shanghai Frontiers Science Center of Molecule Intelligent Syntheses. Meanwhile, we thank the financial support from Shanghai Chinafortune Co., Ltd.

\bibliography{aaai2026}
\clearpage
\section{ADDITIONAL ANALYSIS}
\noindent\textbf{User Study.}
Given the highly subjective nature of texture transfer evaluation, it is challenging to rely solely on objective quantitative metrics for a comprehensive assessment. Therefore, we conducted a user study to evaluate the effectiveness of the proposed method in terms of texture transfer quality. We collected 25 responses from an online questionnaire distributed through social media platforms. The questionnaire consisted of 20 groups of images along with corresponding prompts. Each group included a texture image, an original content image from a random viewpoint, and a rendered image produced by one of the methods under comparison from the same viewpoint. Each rendered image was evaluated in terms of texture alignment and visual quality, with scores ranging from 1 (lowest) to 5 (highest). As shown in Tab.~\ref{tab:user}, our method consistently outperforms the baseline approaches. Specifically, by comparing the differences across various evaluation dimensions, it is evident that users showed a clear preference for our proposed method. This indicates that our approach achieves texture transfer results that better align with human visual perception.

\noindent\textbf{Runtime Cost.}
Runtime efficiency is crucial for 3D assets. Therefore, we evaluate the runtime cost of different methods on the LLFF dataset. As shown in Tab.~\ref{tab:cost}, our method demonstrates superior efficiency in terms of both memory usage and FPS. StyleGaussian incurs higher memory costs due to the use of additional VGG feature parameters during rendering. ARF, Ref-NPR, and SNeRF are NeRF-based methods, which result in slower rendering speeds.

\begin{figure}[t]
    \centering
    \includegraphics[width=1\linewidth]{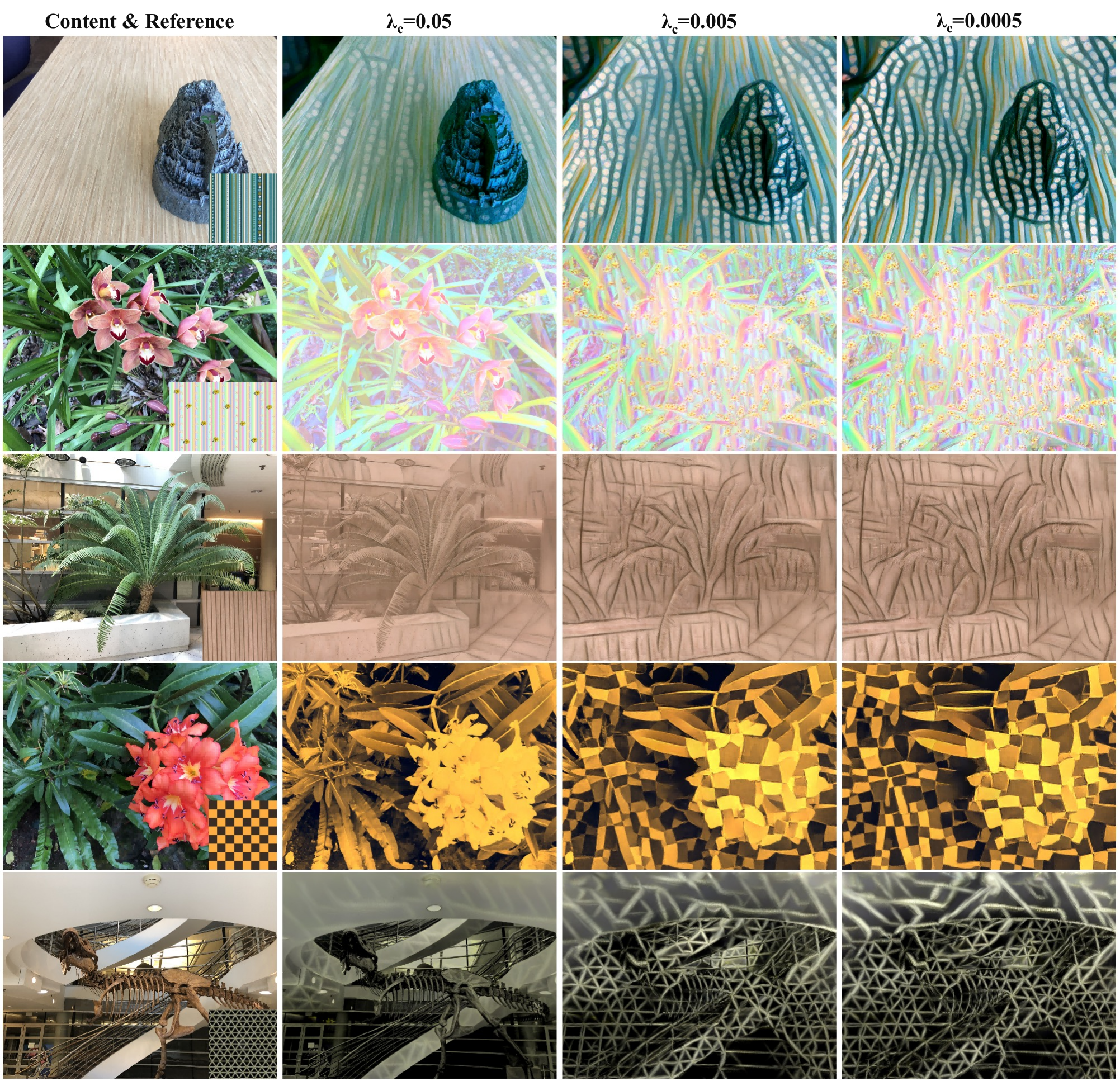}
    \caption{Ablation Study of Content Loss.}
    \label{fig:content}
\end{figure}

\noindent\textbf{Impact of Content Loss.}
As shown in Fig. \ref{fig:content}, the content loss controls the strength of texture transfer, helping preserve the original scene content. However, an excessively large $\lambda_c$ hinders the scene from effectively learning the desired texture appearance.

\section{MORE IMPLEMENTATION DETAILS}
The frequency map is derived from high-frequency energy statistics based on the Discrete Cosine Transform (DCT). Specifically, according to the scale difference between the image and the feature map, the image is divided into 8×8 blocks, and DCT is performed on each block. The frequency density of a given region is defined as the sum of the absolute DCT coefficients within the corresponding block.
For the forward-facing scene dataset LLFF and the 360-degree scene dataset T\&T, we adopt different experimental settings. In the color transfer stage, we perform 400 and 1000 optimization steps for LLFF and T\&T scenes, respectively. 
During the subsequent texture transfer stage, for the LLFF dataset, we set $\{\lambda_{gt},\lambda_c,\lambda_{tv},\lambda_{reg}\}$ = \{2, 0.005, 0.02, 0.1\}. For the T\&T dataset, we reduce parameter $\lambda_c$ to 0.0005.
To simplify the implementation of the geometry preservation branch, our code adopts an alternating optimization strategy between the two branches.

\section{MORE QUALITATIVE EVALUATION}

\begin{figure}[t]
    \centering
    \includegraphics[width=1\linewidth]{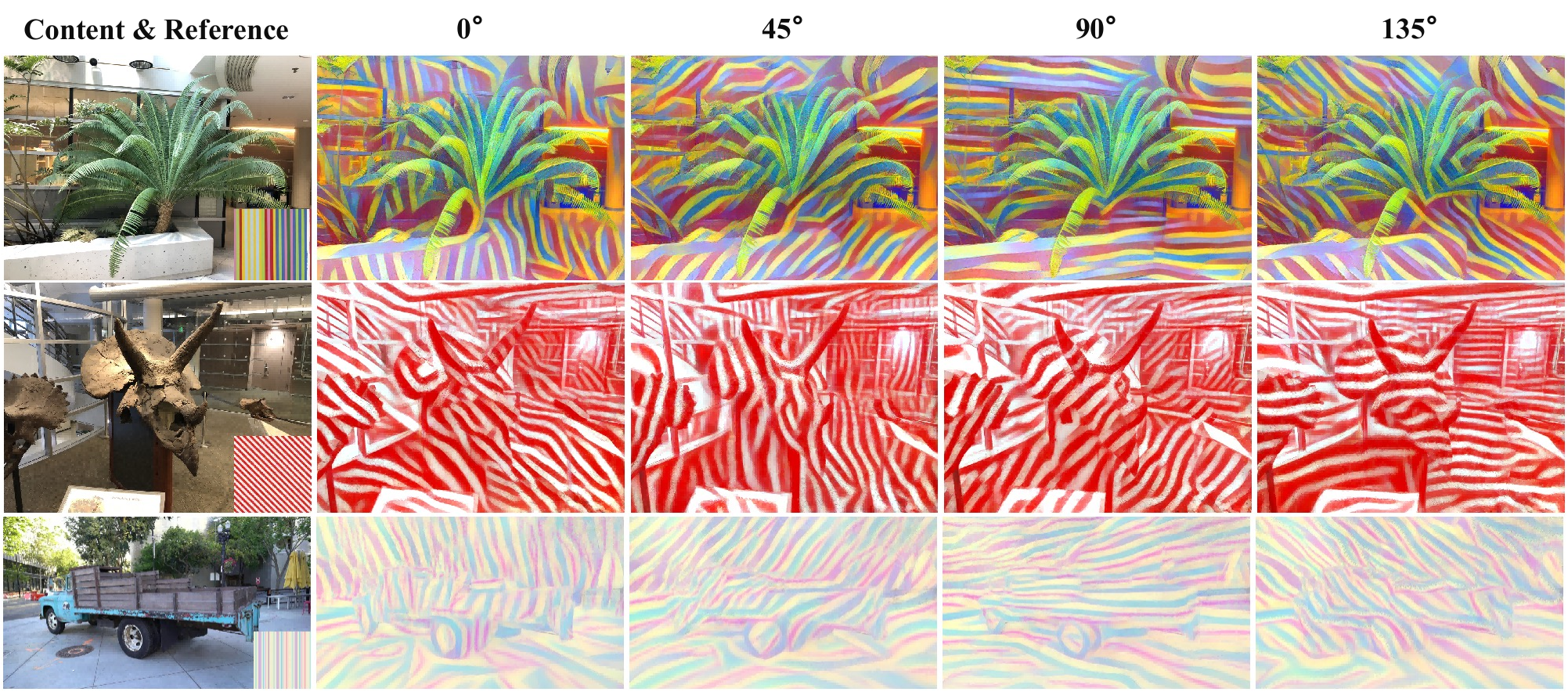}
    \caption{Qualitative Results of Texture Orientation Control.}
    \label{fig:tex}
\end{figure}

\noindent\textbf{Texture Orientation Control.} 
The proposed GT Loss enables the scene to propagate the target texture information based on the prior view. Therefore, by controlling the learnable texture direction in the initial prior view, we can influence the overall appearance learned by the entire scene. This control is achieved by injecting pseudo prior angle information into the first prior view. As shown in Fig. \ref{fig:tex}, by combining the reference texture image and view direction control, we achieve high-quality texture transfer results with controllable texture orientation.

\begin{figure*}[h]
    \centering
    \includegraphics[width=1\linewidth]{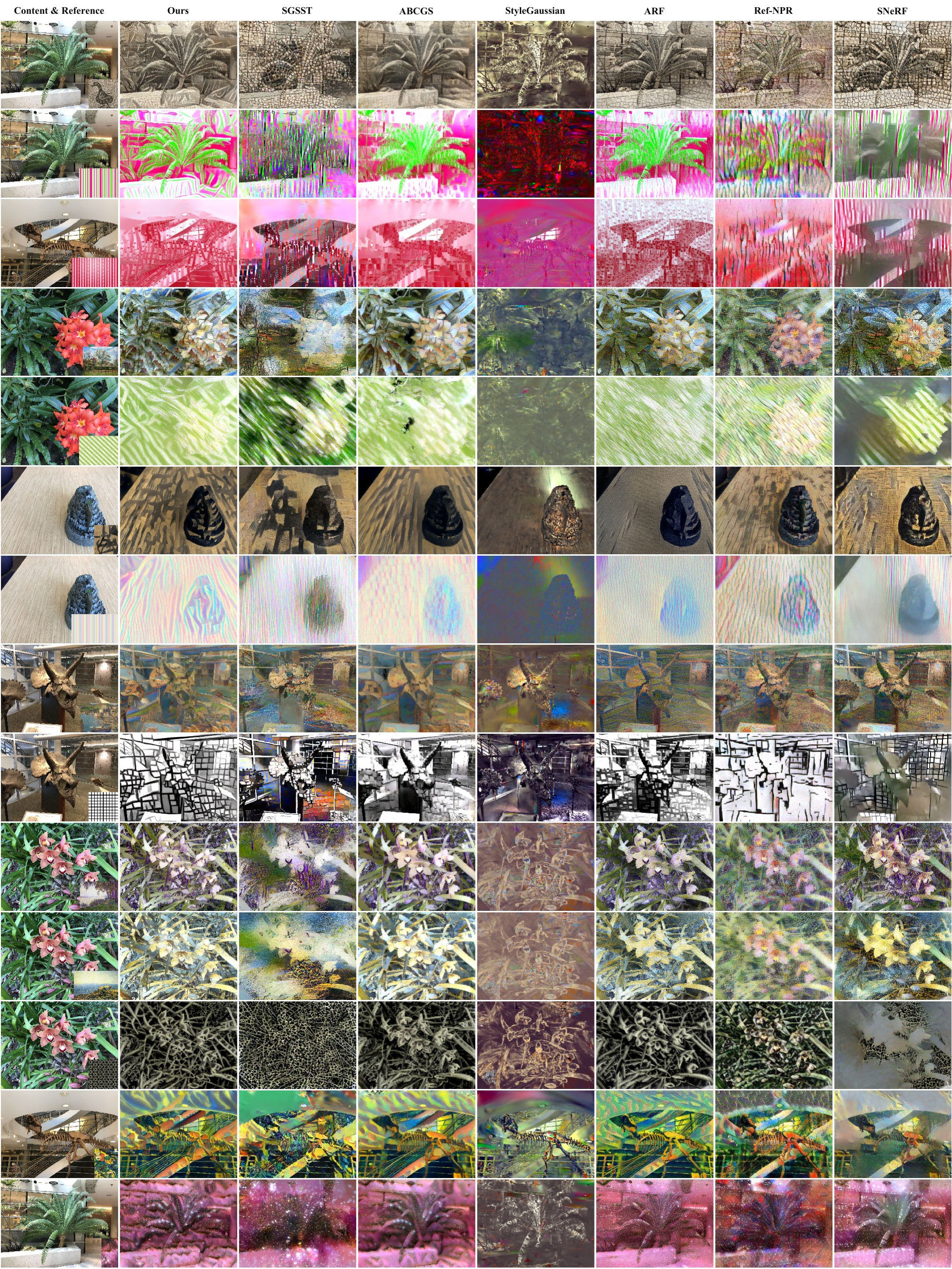}
    \caption{Comparison of Qualitative Results.}
    \label{fig:qual1}
\end{figure*}

\begin{figure*}[h]
    \centering
    \includegraphics[width=1\linewidth]{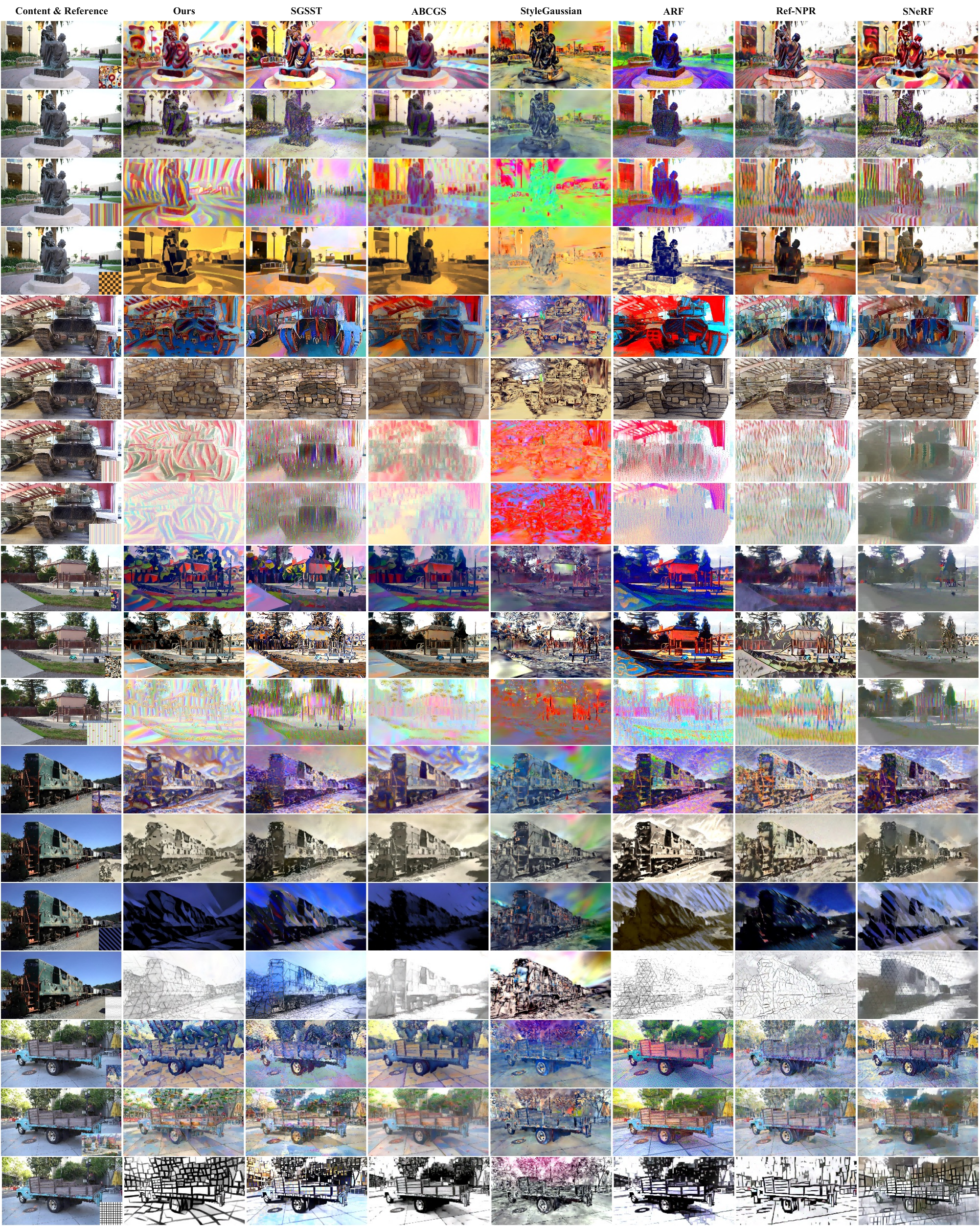}
    \caption{Comparison of Qualitative Results.}
    \label{fig:qual2}
\end{figure*}

\noindent\textbf{More Comparison.}
For more qualitative evaluation, we compare our method with SGSST~\cite{galerne2025sgsst}, ABC-GS~\cite{liu2025abc}, StyleGaussian~\cite{liu2024stylegaussian}, ARF~\cite{zhang2022arf}, Ref-NPR~\cite{zhang2023ref} and SNeRF~\cite{nguyen2022snerf}. Fig. \ref{fig:qual1} and Fig. \ref{fig:qual2} present the qualitative results.

\begin{table}[t]

    \centering
    \setlength{\tabcolsep}{0.5mm}
    \begin{tabular}{ccccc}
        \toprule
                & Ours & SGSST & ABC-GS & ARF   \\
        \midrule
               Texture Alignment($\uparrow$)   & \textbf{4.24}  & 3.16 & 3.04 & 3.04\\
               Visual Quality($\uparrow$) & \textbf{3.92}  & 3.20 & 3.36 & 3.08 \\
        \bottomrule
    \end{tabular}
    \caption{The results of user study.}
    \label{tab:user}
\end{table}

\begin{table*}[t]

    \centering
    \begin{tabular}{cccccccc}
        \toprule
                & Ours & SGSST & ABC-GS & StyleGaussian & ARF & Ref-NPR & SNeRF  \\
        \midrule
               Memory(GB)($\downarrow$)   & \textbf{1.1}  & 1.7 & 1.2 & 6.1 & 1.4 & 1.4 & 1.4\\
               FPS($\uparrow$) & \textbf{151}  & 132 & 133 & 143 & 8 & 7 & 7 \\
        \bottomrule
    \end{tabular}
    \caption{Runtime Cost Comparison on the LLFF Dataset.}
    \label{tab:cost}
\end{table*}

\section{DISCUSSION AND LIMITATION}
Our method primarily consists of three key components: Geometry-aware Texture Transfer Loss (GT$^2$ Loss), Adaptive Fine-grained Control Module (AFCM), and Geometry Preservation Branch (GPB). Among them, the proposed GT$^2$ Loss and AFCM can be seamlessly applied to other representations, such as NeRF, TensoRF, and Plenoxel. Meanwhile, GPB provides a novel perspective for geometry-aware optimization in 3DGS-based appearance editing tasks without ground truth supervision. However, from another perspective, the upper bound of geometric accuracy after texture transfer in our framework depends on the quality of geometry obtained from the original 3DGS optimization, which may be suboptimal. 

\section{FUTURE WORK}
In future work, we plan to focus on two main aspects. First, we aim to enhance the geometry correction stage of GPB by integrating more accurate 3DGS optimization techniques for geometry reconstruction. This improves the quality of scene appearance editing. In addition, we will explore the integration of diffusion models and multimodal large language models (MLLM) with 3D texture and style transfer. Leveraging the rich prior knowledge of multimodal models, we envision a unified framework that supports both text-driven and image-driven style transfer in 3D scenes.

\end{document}